\documentclass[a3paper,11pt]{article}

\usepackage[left=30mm,right=40mm,top=25mm,bottom=20mm, includeheadfoot]{geometry}
\usepackage[rm={tabular, lining},sf={tabular, lining},tt={monowidth, tabular,  
	lining}]{cfr-lm}
\usepackage[onehalfspacing]{setspace}     
\usepackage[utf8]{inputenc}
\usepackage{newunicodechar}
\newunicodechar{ِ}{}
\usepackage{courier}
\usepackage{booktabs}
\usepackage{import}
\usepackage{physics}
\usepackage{amsmath,amssymb,amsthm}
\usepackage{graphicx}
\graphicspath{figures}
\usepackage{tikz}
\usetikzlibrary{arrows,shapes}
\usepackage{acronym}
\usepackage{multirow}
\usepackage{framed}
\usepackage{wrapfig}
\usepackage{xcolor}
\usepackage{ragged2e}
\usepackage{slashed}
\usepackage{dsfont}
\usepackage{mathtools}
\usepackage{caption}
\usepackage{subfig}
\usepackage{multirow}
\usepackage[skins]{tcolorbox}
\usepackage{cancel}
\usepackage{float}
\usepackage{chapterbib,rotating}
\usepackage[linewidth=2pt]{mdframed}
\usepackage{tocbibind}
\usepackage[nonamebreak,numbers]{natbib}
\usepackage{lscape}
\usepackage{natbib}
\usepackage{subfiles}
\usepackage{ifthen}
\usepackage{url}
\usepackage{ulem}

\begin{document}
	
	\title{\textbf{Photonic Neural Networks: A Compact Review}}
	
	\author{\textbf{Mohammad Ahmadi}\\
		Laser and Plasma Research Institute, University of Shahid Beheshti, Tehran, Iran\\
		mohammad.ahmadi1@mail.sbu.ac.ir\\
		\and \textbf{Hamidreza Bolhasani}\\ Department of Computer Engineering,\\ Science and Research Branch, Islamic Azad University, Tehran, Iran\\
		hamidreza.bolhasani@srbiau.ac.ir}
	
	\date{}
	\maketitle
	
	\vspace{-0.5cm}
	\section*{Abstract}
	\noindent
	
	\textbf{It has long been known that photonic science and especially photonic communications can raise the speed of technologies and manufacturing. More recently, photonic science has also been interested in its capabilities to implement low-precision linear operations, such as matrix multiplications, quickly and efficiently. For a long time, most scientists taught that Electronics is the end of science but after many years and about 35 years ago it has been understood that electronics do not answer alone and should have a new science. Today we face modern ways and instruments for doing tasks as soon as possible in proportion to many decades before. The velocity of progress in science is very fast. All our progress in science area is dependent on modern knowledge about new methods. In this research, we want to review the concept of a photonic neural network. For this research, articles were selected from 2015 to the 2023 year. These articles noticed three principles: 1- Experimental concepts, 2- Theoretical concepts, and, finally 3- Mathematic concepts. One of the topics that are very valid and also new, is simulation. We used to work with simulation in some parts of this research. First, briefly, we start by introducing photonics and neural networks. In the second we explain the advantages and disadvantages of a combination of both in the science world and industries and technologies about them. Also, we are talking about the achievements of a thin modern science. Third, we try to introduce some important and valid parameters in neural networks. In this manner, we use many mathematical tools in some portions of this article.}
	\\\\
	\hspace{2cm}\textbf{\textit{Keywords}: Photonics, Neural Networks, Electronics, Photonic Neural Network, Convolutional Neural Network, Optical Neural Network.}
	
	\section{Introduction}	
	\label{sectin.Introduction}
	\textbf{Photonics} is one of the newest sciences. Photonics is the study of light and interaction with other particles or materials. May we say photonic science includes and is based on 4 main topics: 1-ray optics, 2-wave optics, 3-Electromagnetic optics, and 4- Quantum optics. quantum optics is the best complete theory about all of the physical concepts in the world. \textit{Photonics} is like a bridge between Optics or Physics and the other sciences such as Electronics or Chemistry and Biology and so more. You should be very careful in the definition of photonic and optic. The optic is not photonic and vice versa but the optic is a subpart and the most important part of photonic. We in photonics are faced with interaction among waves and collision between particles - especially Photons - with the matter. Also, we concentrate on other applications of it, for example, modulation and modifying waves. Photonics has explained many topics such as Fiber, Imaging, Laser Processing, Tera Hertz, etc. It started in the 1960s and opened many doors of science concepts between people. But applications of photonics are very bright when we combine them with computers and neural networks.
	
	\noindent
	\textbf{Neural networks} is a modern method of knowledge that is very applicable because it is very active in treatment ways and computing systems. Or in a more accurate definition we could say, is a series of algorithms that endeavors to recognize underlying relationships in a set of data through a process that mimics the way the human brain operates. Neural networks help us to be a classifier. In 1943, neurophysiologist Warren McCulloch and mathematician Walter Pitts wrote a paper on how neurons might work and we could say Neural Networks started. Neural networks can adapt to changing input; so the network can generate the best possible result without needing to redesign the output criteria. It is related to Neurons and it has very important rules and interaction with other sciences. Through it, we can transform our information and process them very quickly. Neural Networks or generally AI(Artificial intelligence) based on two methods: 1-\textit{training} in which the models learn to know sets of data and 2-\textit{inference} in which try to understand the results and conclusions of the input data. Figure1 had been tried to explain a simple schematic of a Neural Network process. This picture consists of inputs, functions, and output. The first layer is input and function ($ w $) operates it and then by passing from two hidden layers ($ b_{3} and b_{4} $) achieves the output layer ($b_{1}$) and then final eigenvalue ($ a $). \\
	For simple showing been entered some inputs and had been done many multiplications of math on them and finally, we received outputs.\\
	Consequently, photonic science
	is a perfect alternative for electronic [ 28,29 ].\\
	At first, much manufacturing was bulky and made from insufficient technology and Neural Network concepts. Recently
	it has changed, first of all, because RC enabled the reduction of analog electronic and photonic RNN.
	[ 30-34 ].\\
	In addition, integrated photonic platforms also have been completed and today they are practical [ 35 ].
	Yet, Neural Networks
	consist of numerous photonic nonlinear nodes. So it has been demonstrated
	only in delay systems [ 36-39 ].\\
	Due to the time multiplexing, delay system Neural Networks require such a subsidiary foundation [6].
	\\
	Photonic technologies also are widely applied and have validity in our daily lives.
	Integrated photonics ar first [40], and after that indium phosphide–based technologies [41], emerged for optical communication, metro, and also for short-range links. The advantages of optics that are very important: 1- larger bandwidth-distance crops,
	2- the massive parallelism, 3- low propagation loss, etc. This can be transmitted through optical ﬁbers [42,9]. Also in Figure 4, we displayed a brief history and prediction of the future of Photonic Neural Networks.

	\subsection*{Research Methodology}
	In this review article, we explored many articles from strong scientists and experts around the world. We tried to apply the main topics and research among many of the articles that we used. These articles some of them were very bold and perfect. This sentence means some articles such as the survey article have been applied many times in this exploration and they were very useful. Well actually as you saw most of these articles were published generally from 2017 to 2022. After this introduction, we explain an overview of a photonic neural network at \textit{Application} part. Then we extend some fundamental concepts and finally, we will discuss the advantages and disadvantages.\\
	Table 1 provides some main information about the reviewed articles
	contained perfect information about opportunities and applications of Photonic Neural Networks.

	In this research, we want to answer the following important questions:\\\\
	\textbf{Q1}:\;\textit {Why Photonic Neural Networks is important?}\\
	\textbf{Q2}:\;\textit{Although Electronic has progressed at many parts of industrial work or other things, what is the location of Photonics and what are the challenges between Electronics and Photonics?}\\
	\textbf{Q3}: \textit{What are the most important challenges of Photonic Neural Network?}\\
	\textbf{Q4}: \textit{What are the acheivments of Photonic Neural Networks?}\\
	
	\begin{table}[htbp]
		\centering
		\caption{Major methods of studied research included PNNs}
		\begin{tabular}{ccccccccccccccccccccccccccccccccccccccc}
			
			\toprule
			
			Reference & Publication
			year  & Main context
			\\
			\midrule
			Kun Liao et al.& 2023&  Opportunities and Challenges in Photonic Neural Networks\\
			\midrule
			Cansu Demirkiran et al.[74]	& 2023   &   An Electro-Photonic System for Speeding up DNNs\\
			\midrule
			YUN-JHU Lee et al.& 2023 & Introduction of Photonic spiking neural networks (PSNNs)\\
			\midrule
			Sanmitra Banerjee et al.& 2023 & Mach-Zehnder Interferometers in coherent Integrated Photonic
			Neural Networks (IPNNs)\\
			\bottomrule
		\end{tabular}
		\label{default}
	\end{table}
	
	\begin{figure}[b!]
		\begin{center}
			\includegraphics[width = 25cm , height = 16cm]{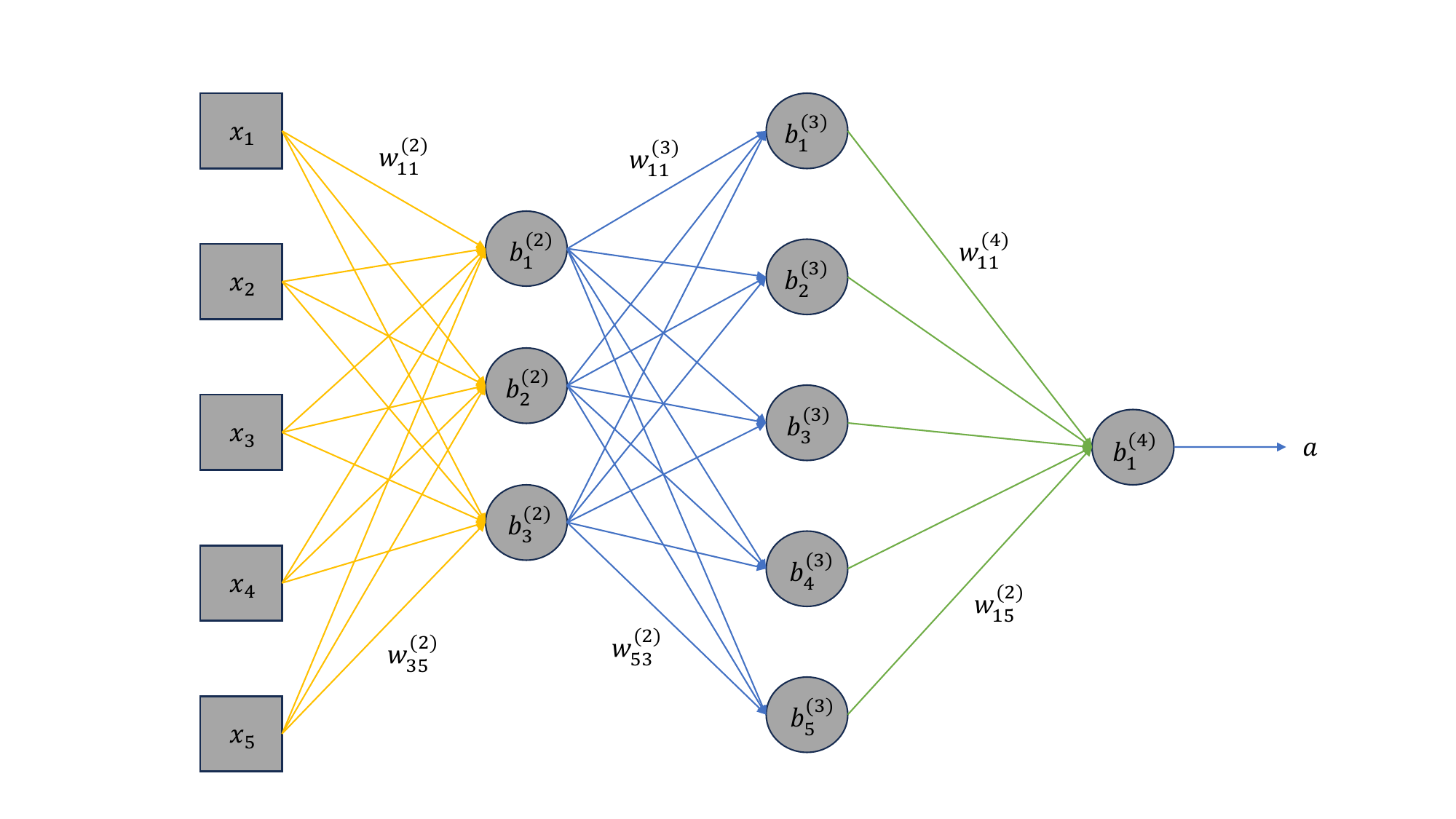}
			\caption{Schematic for Neural Network includes First Colomn: Input Layers, Second And Third Colomn: Hidden Layers, Final part: Output Layer.}
		\end{center}
	\end{figure}

	\textbf{Q1} : Why Photonic Neural Networks is important?

	\section{Fundamental Concepts}
	
	\subsection{Roadmap on material-function mapping}
	
	some parts of this topic explained the differences between photonic and memristor. (pros and cons). At it, we can see the perception model, which is the first neural network.\\
	Strength and pain points for photonic information processing:\\
	First, we are going to say some benefits: 1- Real-time. 2- Parallelism, etc. \\
	And now we will say the disadvantages: 1- Efficient memory. 2- Functionality.

	\begin{figure}
		\vspace{-1.2cm}
		\begin{center}
			\includegraphics[width = 18cm , height = 12cm]{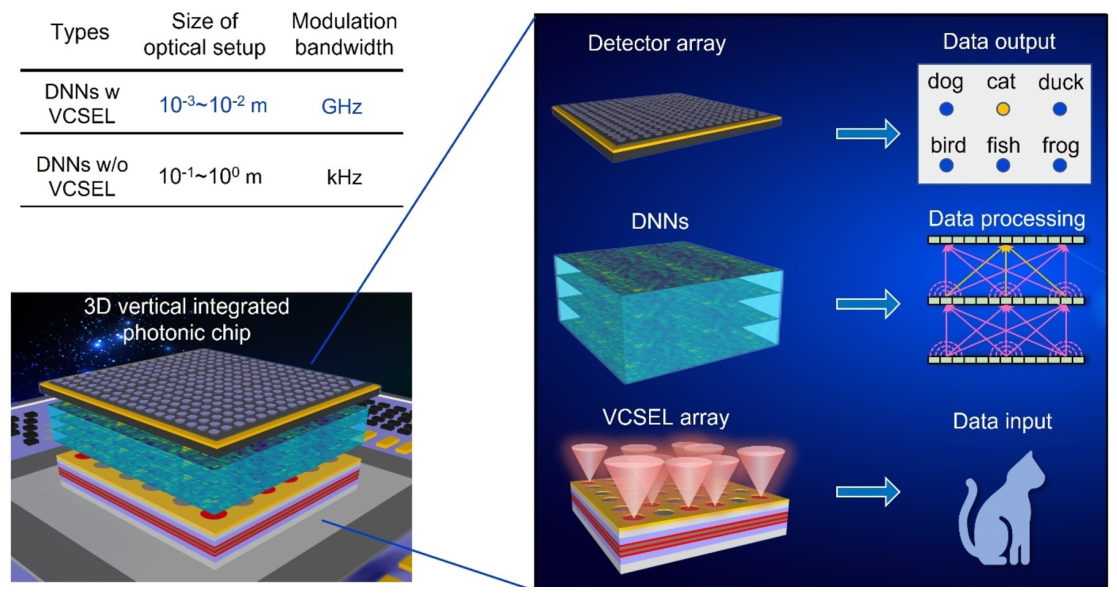}
			
			\caption{Application of DNNs (Deep Neural Networks) for VCSEL (vertical-cavity surface-emitting laser) [71].}
		\end{center}
	\end{figure}

	\subsection{Integrated Photonics}
	This subject provides a good manner to discover huge-scale optical networks on chips. one of the applications of photonics is less distortion. Also, the modern technology of it, standardization of silicon photonic integrated circuits (PIC) has led to the proliferation of shorter-distance photonic links.\\

	Photonic devices have become very dense compared to now.\\
	Because the concept of training is very bold and important for neural networks, Integrated Photonic devices are necessary. As previously said, IPNNs play a great role in phase modulation. It has been used in many number of MZI [73].
	
	\subsection{Neuromorphic Photonics} 
	It is useful to work by matrix-vector multiplication (MVM).\\
	Now again we want to talk about some information related to silicon. Silicon photonic modulator neuron is one of the topics at neuromorphic. This field exists in pivotal and unexplored regimes of machine intelligence. We have to say neuromorphic photonic systems can calculate and operate about 8 orders faster than electronic samples.\\
	Then we are going to define a device: The modulator neuron is an optical-to-electrical-to-optical device consisting of two balanced photodetectors. (PDs)\\
	Characterization: There are three independent quantities:\\
	1- Heater current bias ($I_{h}$)\\
	2- Modulator current bias ($I_{b}$)\\
	3- Optical power into the $IN+$ and $IN-$ parts.($P_{+}$ ,$P_{-}$ ,...)

	\subsection{Convolutional Neural Network (CNN)}
	
	Convolutional Neural Network (CNN) is one the important parts of AI systems. This is a powerful series of artificial neural networks. The application of CNN is at the recognition of handwritten digit images perfectly and strongly. We need the shaped microvomb's optical spectrum and the corresponding kernel weights.\\ Convolutional Neural Network is a great kind of deep learning. In 1959, Hubel and Wiesel found that cells in the animal
	visual cortex are responsible for detecting light in receptive fields. And then Kunihiko
	Fukushima could propose the recognition in 1980, which could be regarded as the predecessor of CNN. In
	1990, LeCun published a seminal paper establishing the modern framework of CNN and later
	improved it [20].
	
	\begin{figure}[b!]
		\begin{center}
			\includegraphics[width = 15cm , height = 12cm]{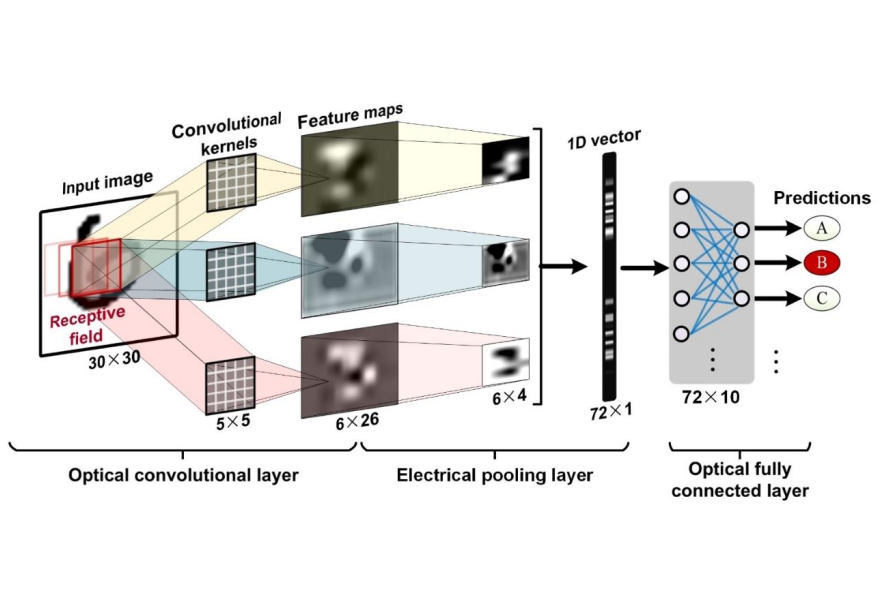}
			\vspace{-1.8cm}
			\caption {A schematic for \textbf{Convolution Neural Network (CNN)}[11].}
		\end{center}
	\end{figure}

	\subsection{Optical Neural Networks (ONNs)}
	Optical neural networks can process information in free space and an integrated platform. Over many years, the system of integrated circuits has been restricted by Moore’s law [21-27]. Today, Optical Neural Networks (ONNs) in photonic devices are higher efficient and have lower power consumption than electronic platforms ANN in deep learning consists of research attention such as image processing, self-driving vehicles, games, and robots[2]!\\
	Photonic devices also incorporate those
	nonvolatile PCMs thus can:\\
	1- realize optical memories, 2- perform
	in-memory computing simply[64–66].
	increasing for these phase-change photonic devices in a scalable
	network has been
	designed and developed[66–69,15].

	\subsection{Three-dimentional waveguide interconnects}
	
	We had some problems or limitations with two-dimensional work. For example, limitation in size by lithography due to disadvantages of scaling. One good manner to overcome many bottlenecks in NN communication is to discover integrated circuits holding to an NN's difficult topology.\\
	3D interconnects of the photonic waveguide are very useful in 3D commercial use direct-laser writing system called (Photonic Professional GT). Also, commercial NN is heavily employed by Google features.\\
	We realized large-scale 3D photonic interconnects. Waveguides have a diameter
	of 1.2 $ \mu $m that could create two-photon polymerization. By use of this strategy,
	we discovered routing topologies at 3 dimensions. These
	methods were oriented mostly toward applications in Neural Networks [19].
	
	\subsection{Large-scale Optical Neural Networks}
	
	One of the important features and instruments that name is $neural-network\; accelerator$. It is useful for the implementation of a single layer. In this location, we do many acts of math, for example, multiplier integrator [14].
	
	\begin{equation}
		E_{tot}=(mh+nk)E_{in}+(mn)E_{out}
	\end{equation}
	\begin{equation}
		E_{MAC}=\left(\dfrac{1}{n}+\dfrac{1}{m}\right)E_{in}+\dfrac{1}{k}E_{out}\;\;\;\;\;\;\;\;\;\; 
	\end{equation}
	In this equation, $n, m, k$ are matrix dimensions. And in there $E_{in}$ and $E_{out}$ are related to transmitting and receiving the Energy. Like this equation, we can rewrite for MAC(Multiply And Accumulate):\\
	\begin{equation}
		E_{MAC}=\left(\dfrac{1}{C^{'}}+\dfrac{1}{W^{'}H^{'}}\right)E_{in}+\dfrac{1}{K_{x}K_{y}C}E_{out}
	\end{equation}
	We use this method to compute the required matrix products optoelectronically without the need for an all-optical nonlinearity. Computing used by Monte Carlo simulations. 
	Also, this feature is a new kind of photonic accelerator that is scalable to values of $N$ that are larger than  $10^{6}$  networks. And also can be operated at high (Gigahertz) velocities and consumes very low energies. For coherent matrix we will have relationship between $ 10(th)$ and $(k+1)th $ layers[14]:
	\begin{equation}
		x_{i}^{k+1}=f(\sum_{j}A^{(k)}_{ij}x^{(k)}_{ij})\;\;\Longrightarrow\;\;f\equiv activate \; function
	\end{equation}

	\section{Progress and Development of Photonic Neural Networks}
	PHOTONICS has been realized for its important role in communication systems. For example, fiber optic links form the world’s telecommunications. So we could say a photonic
	waveguide, with sizes ranging about (80 $ \mu $m)
	, can carry information very easily per second without depending on distance [17].\\
	Today machine learning science is developed for many tasks and it is used in
	a large number of processing and many other applications such as easy calculation, image information
	classification, speech recognition, language translation, decision making, web searches,
	etc. Artificial Neural Networks or ANNs also are useful for
	processing data series, combining and analyzing information quickly [5].
	Electronic pieces of information are very necessary but electronics face a main impasse: data transfer between processors and memories is constrained by unavoidable bandwidth limitations.\\
	One another application of photonic is at the L3 cavity:\\
	Our purpose is to find ways to optimize and increase the Q-factor in 3L. The cost of this work is very much so we have to use NN or Neural Network. This work is done in several steps:\\
	1. Preparing phase, 2. Learning phase, 3. Structure search phase and, finally 4. Validation and update phases[8].\\
	Another important use of photonics well actually as you saw, is Deep Neural networks:
	They are used in a range of applications such as image and video processing, diagnostic medical imaging, speech recognition, and conversational AI \\
	Deep learning ways have been tried to obtain non-linear operators
	that transform the representation from the previous levels into a higher one.\\
	We face an important concept called optical response. The optical response was predictable. By using deep learning neural networks could predict the optical response of such information.
	Also, we used Bernoulli probably classifier. (for hole location) and we used the Gaussian log-likelihood loss function[18]:
	\begin{equation}
		-log \;P(Y|X,\vec{W})=\dfrac{1}{K}\sum_{n}^{K}(\dfrac{1}{2}log(2\pi \sigma^{2})+\dfrac{1}{2\sigma^{2}}(y_{n}-\vec{W}^{T}x_{n})^{2})
	\end{equation}
	In the above equation, $P(Y|X,\vec{W})$ is a function that is related to probabilities and $\vec{W}$ is related to model parameters. Also, $K$ is the number of training data. 
	Also, we explain the process of executing a given
	DNN on ADEPT. The compiler (Figure 6) takes a DNN model as input and compiles it to generate a program speciﬁed via
	a graph on tensor types. It has a cost-model-based partitioner
	that creates a directed acyclic graph (DAG) with 3 sets of
	nodes: operations to execute on the host CPU (e.g., input pre-
	and post-processing, etc.); operations to execute on ADEPT
	(e.g., ML operations such as GEMM, convolutions, non-linear
	operations, etc.); and a set of driver nodes for moving data
	between the host and ADEPT[3].
	Emerging photonic neural networks increased computing speed by 2-3 orders of magnitude. Two advanced features were: weighted sum and NL activation, The Absence of NL(non-linear)  in optical systems has severely limited its usefulness in deep learning computing.
	All work and activities goal to progress both speed and energy efficiency. One of these achievements is ANN or Artificial Neural Networks.
	Then we have to discuss biology. Biological neural networks consist of a huge number of neurons. In this work, the optimization variables for all ONNs are controlled by a kind of perfect velocity [12]:
	\begin{equation}
		V_{i}^{k+1} = WV_{i}^{k}+c_{1}r_{1}(pb_{i}^{k}-X_{i}^{k})+c_{2}r_{2}(gb_{k}^{d}-X_{i}^{k})
	\end{equation} 
	That is the equation:  \\
	\begin{equation*}
		c_{1}=c_{2}=1.49445
	\end{equation*}
	Are acceleration constants.\\
	For designing integrated photonic devices have to say 
	integrated photonics manners have become powerful tools for processing at classical and
	quantum information. Integrated photonics is used for
	1- optical interconnects[ 43 ], 2- signaling [ 44 ], and finally photonic processing [ 45 ].About Processing
	quantum information have to say it also indicates great agreement [ 46, 47 ], with discovered 
	advantages at quantum communications [ 48 ] Also
	for quantum simulation [ 49, 50 ].
	Designing integrated photonics circuits, however, remains a major impasse
	[ 51 ]. However current designs are complicated by computational capability [52,10].
	Photonic physics science exhibits properties distinct from those
	of electronics in terms of multiplexing, energy dissipation,
	and cross-talk. These properties are favorable for dense,
	high-bandwidth interconnects [53] in addition to conﬁgurable analog signal processing [54–56]. As a result,
	neuromorphic photonic systems also could operate in 6–8 orders-
	of magnitude faster than neuromorphic electronics [63]
	with the possibility of higher energy effects than others [57]. Optical neural binds based on ﬁeld progress in 1- free space [58, 59],2- holograms [60, 61], and also ﬁber [62] have
	been shown but were not widely corresponded, in part because they could not be integrated on a chip[13,16].
	
	\noindent After that, we will pay attention to the Challenges and Benefits of photonic neural networks. Then we try to summarize the main actions that have recently been done in photonic neural networks.\\
	In Table 4 this information has been explained.
	
	\noindent
	In a complete photonic neural network, you would need to model and connect multiple such components to perform optical computations for neural network tasks. Additionally, you may need to implement backpropagation or other training algorithms specific to your neural network architecture.\\
	From 1997 still now, knowledge of AI has progressed on a very large scale. We will demonstrate it in the next part of the article.\\
	Another strong application in photonic neural networks is at photonic crystal waveguides (PhCWs). Photonic crystals have a very large-scale application and they exist in 1D, 2D, and 3D [71].

	\subsection{Application in Integrated Photonics}
	
	In some studies that have been researched the number of simulations for the process of Integrated Photonics is too much and also is weak in the speed. So using a Deep Neural Network could redesign these Integrated Photonics inversely. This work aims to reach the great Power Splitter based on Integrated photonics.\\
	As we know the emergence of IPNN(Integrated Photonics Neural Networks) could progress many computational activities and speed up them. In the following text, we explore some of these applications:\\\\
	\textbf{1. On-Chip Waveguide-Based IPPNs}\\
	In this manner, we could use MZI(Mach-Zehnder Interferometer) and the concept of the phase shifter, converting phase modulation into an interesting modulation! However one of the important challenges in this way is the limitation of optical instruments and concepts. For example, in neural networks, we don't see bounded conditions but in optics, we face Energy conservation concepts.\\
	
	\begin{table}[htbp]
		\centering
		\caption{Acheivments related to On-Chip Waveguide-Based IPNNs.}
		\begin{tabular}{ccccccccccccccccccccccccccccccccccccccc}
			
			\toprule
			
			Refrence & Publication
			year  & Acheivment
			\\
			\midrule
			Zhou et al.[75]& 2016&  A little precision optical switch\\
			\midrule
			Soljačić et al.[75]& 2017   &   Perfect ability for computation at optical feed-forward neural networks\\
			\midrule
			Dong et al.[75]& 2020 & Making silicon photonic signal processor\\
			\midrule
			Hu et al.[75]& 2021 & Advanced photonic
			neural network with 27 channels that are \\\;&\;&useful in Convolutional Neural Networks(CNNs)\\
			\midrule
			Hu et al.[75]& 2022 & Producing five-layer IPNNs for adoption in nonlinear process.\\
			\bottomrule
		\end{tabular}
		\label{default}
	\end{table}

	\begin{table}[htbp]
		\centering
		\caption{Acheivments related to On-Chip ِDielectric Metasurface-Based IPNNs}
		\begin{tabular}{ccccccccccccccccccccccccccccccccccccccc}
			
			\toprule
			
			Refrence & Publication
			year  & Acheivment
			\\
			\midrule
			Li et al.[75]& 2021&  Important application for control the switching of special modes and \\ \;&\;& demonstrating a metasurface waveguide made a good converter.\\
			\midrule
			Cheng et al.[75]& 2021   &  Using several(three) silicon slots as a single neuron that is\\\;&\;& a great structure in large-scale than CMOS process\\
			\midrule
			Gu et al.[75]& 2022 & Like previous article evidence 1D metasurface that set on SOI substrate\\
			\midrule
			Liu et al.[75]& 2022 & Knowing a silicon based on a kind of technology\\\;&\;& consist of many number of MZI for Fourier transform\\
			\bottomrule
		\end{tabular}
		\label{default}
	\end{table}

	\noindent\textbf{2. On-Chip Dielectric Metasurface-Based IPPNs}\\
	By using this method we can replace traditional MZI with a new optical metasurface. Metasurface materials are arrays of antennae that can control light-matter interaction in both linear and nonlinear regimes at the nanoscale. This way is compatible with on-chip CMOS processes. \\\\
	\noindent
	\textbf{3. Integrated Photonic Spiking Neural Networks}\\
	\noindent
	This process is an efficient way between other methods. Some features of it, include:\\
	1- Spiking neuron produced in subthreshold pulses.\\
	2- Also spiking neuron is very powerful to able to excite the next neuron.\\
	Two main typical features are stated above. According to research in years 2021 and 2019 for controlling charge transformation between graphene and the $MoS{_{2}}$ layer. And in 2019 a kind of photoelectronic synaptic made based on a 2D $MoS_{2}$ phototransistor [76].\\
	
	\subsection{Application in IQP (Integrated Quantum Photonic)}
	This branch of science is interested in focusing on the application of quantized light-matter interaction. One of the very useful achievements is Hybrid IQP devices. Another progress is about single photon sources and making it, because it is very important in quantum cryptography. But as stated, in the next decade this technology could progress and Photonic Machine Learning will develop. These three areas are connected: Quantum Photonics, Topological Photonics, and Machine Learning Photonic [1].

	\noindent It may be useful that we display some very high resolution and very applicable designs (Application or Benefit) in a photonic neural network in table 5.\\\\

	\begin{center}
		\begin{tabular}{|p{6cm}|p{3.5cm}|p{10cm}|}
			\hline
			\begin{center}
				\textbf{References}
			\end{center} & \begin{center}
				\textbf{Publication year}
			\end{center} & \begin{center}
				\textbf{Main idea}
			\end{center} \\
			\hline  
			\vspace{0.2cm}	\begin{center}
				Mario et al.[4]
			\end{center} &\vspace{0.2cm} \begin{center}
				2019
			\end{center} &\vspace{-0.2cm} Providing a new manner to pave the way for emerging
			photonic-electronic neural networks devices by finding a detailed knowledge in the single nodes \vspace{0.2cm}\\
			\hline 
			\begin{center}
				Mohammad et al.[18]
			\end{center} & \begin{center}
				2019
			\end{center} &\vspace{-0.2cm} Using a deep neural network to introduce a method for predicting optical response \\
			\hline  
			\begin{center}
				LORENZO  et al.[5]
			\end{center} & \begin{center}
				2019
			\end{center} &\vspace{-0.2cm} Reviewing for achievement at  photonic neural network\\
			\hline    
			\begin{center}
				TIAN et al.[12]
			\end{center}& \begin{center}
				2019
			\end{center} &\vspace{-0.2cm} Proposing a strategy for learning neural networks based on the neuron to design and train ONN \\
			\hline
			\vspace{0.4cm}\begin{center}
				ALEC et al.[10]
			\end{center}&\vspace{0.4cm} \begin{center}
				2019
			\end{center} &\vspace{-0.2cm}  Training ANNs to model both strip waveguides and chirped Bragg gratings
			using a small number of simple input and output parameters relevant to designers of integrated
			photonic circuits \vspace{0.2cm}\\
			\hline
			\vspace{0.3cm}		\begin{center}
				Xingyuan et al.[11]
			\end{center}&\vspace{0.3cm} \begin{center}
				2020
			\end{center} &\vspace{-0.2cm} Optical neural networks ways based on high speed and low energy consumption  \vspace{0.2cm}\\
			\hline
			\vspace{0.4cm}\begin{center}
				Cansu et al.[3]
			\end{center}& \vspace{0.4cm}\begin{center}
				2021
			\end{center} &\vspace{-0.2cm}  Introducing a complete electronic-
			a photonic system that combines an electronic host processor and DRAM alongside an electronic photonic accelerator
			(called ADEPT) \vspace{0.2cm}\\
			\hline
			\vspace{0.4cm}\begin{center}
				Sunil Pai et al.[72]
			\end{center}& \vspace{0.4cm}\begin{center}
				2023
			\end{center} &\vspace{-0.2cm}  Introducing a novel method for power monitoring in a feedforward photonic neural network by using two detectors.
			\vspace{0.2cm}\\
			\hline
			\vspace{0.4cm}\begin{center}
				Zhigang Chen et al.[1]
			\end{center}& \vspace{0.4cm}\begin{center}
				2021
			\end{center} &\vspace{-0.2cm} Explaining Quantum Photonic neural networks and land space of this novel field. Also, IQP has some achievements that are very valid for the future of Quantum Computing.
			\vspace{0.2cm}\\
			\hline
		\end{tabular} 
		\captionof{table}{Main studied articles}
	\end{center}
	\vspace{0.5cm}
	\newpage
	\noindent
	In the above table, we tried to explain a summary of the main idea of applicable articles from 2018 still 2023. Then in Table 2, we tried to explain the benefit or application of the most important article in recent years.\\
	
	\textbf{Q4}: \textit{What are the acheivments of Photonic Neural Networks?}\\

	\begin{center}
		\begin{tabular}{|p{7cm}|p{3.5cm}|p{10cm}|}
			\hline
			\begin{center}
				\textbf{References}
			\end{center} & \begin{center}
				\textbf{Publication year}
			\end{center} & \begin{center}
				\textbf{Benefits\\(Novel Application)}
			\end{center} \\
			\hline  
			\vspace{0.2cm}	\begin{center}
				MARIO MISCUGLIO et al.[70]
			\end{center} &\vspace{0.2cm} \begin{center}
				2018
			\end{center} &\vspace{-0.2cm}1- Make new devices based on LMI (Light-Matter Interaction) in hybrid systems. This device used 2 oscillators and applied Quantum Dots (QD) between two gold nanoparticles. 2- Using AONN (All Optical Neural Network) to increase the speed of computational activity in comparison with the Electrical model. \vspace{0.2cm}\\
			\hline 
			\begin{center}
				Johnny Moughames et al.[19]
			\end{center} & \begin{center}
				2020
			\end{center} &\vspace{-0.2cm} Invention of 3D waveguide interconnects have evolution in photonic devices. For imaging and scanning had been used SEM (Scanning Electron Microscope). \\
			\hline  
			\begin{center}
				Mitchell A. Nahmias et al.[17]
			\end{center} & \begin{center}
				2020
			\end{center} &\vspace{-0.2cm} This article paid attention to the comparison of Energy consumption between electronic and photonic devices by two perfect plots. As before we discussed consumption in electronic devices versus vector sizes and bits of precision. We can see the differences between these two kind of devices. \\
			\hline    
			\begin{center}
				Pascal Stark et al.[9]
			\end{center}& \begin{center}
				2020
			\end{center} &\vspace{-0.2cm} Two important opportunities were discussed: 1- using integrated photonic into the artificial neural network 'Training'. 2- This feature is capable of neural network inference that in it explored about Mach-Zehnder interferometer.  \\
			\hline
			\vspace{0.4cm}\begin{center}
				Min Gu et al.[71]
			\end{center}&\vspace{0.4cm} \begin{center}
				2023
			\end{center} &\vspace{-0.2cm}  VCSEL (vertical-cavity surface-emitting laser) technique is very useful in photonic neural network devices. VCSEL is a kind of semiconductor that is very important in 3D integrated photonics. some of the benefits are: 1-low threshold, 2-high modulation bandwidth of more than 30 Ghz,  3- Optical encryption, 4- Image classification\vspace{0.2cm}\\
			\hline
		\end{tabular}
		\captionof{table}{Some of the improvements in photonic neural networks}
	\end{center}
	
	\begin{figure}[b!]
		\begin{center}
			\includegraphics[width = 23cm , height = 14cm]{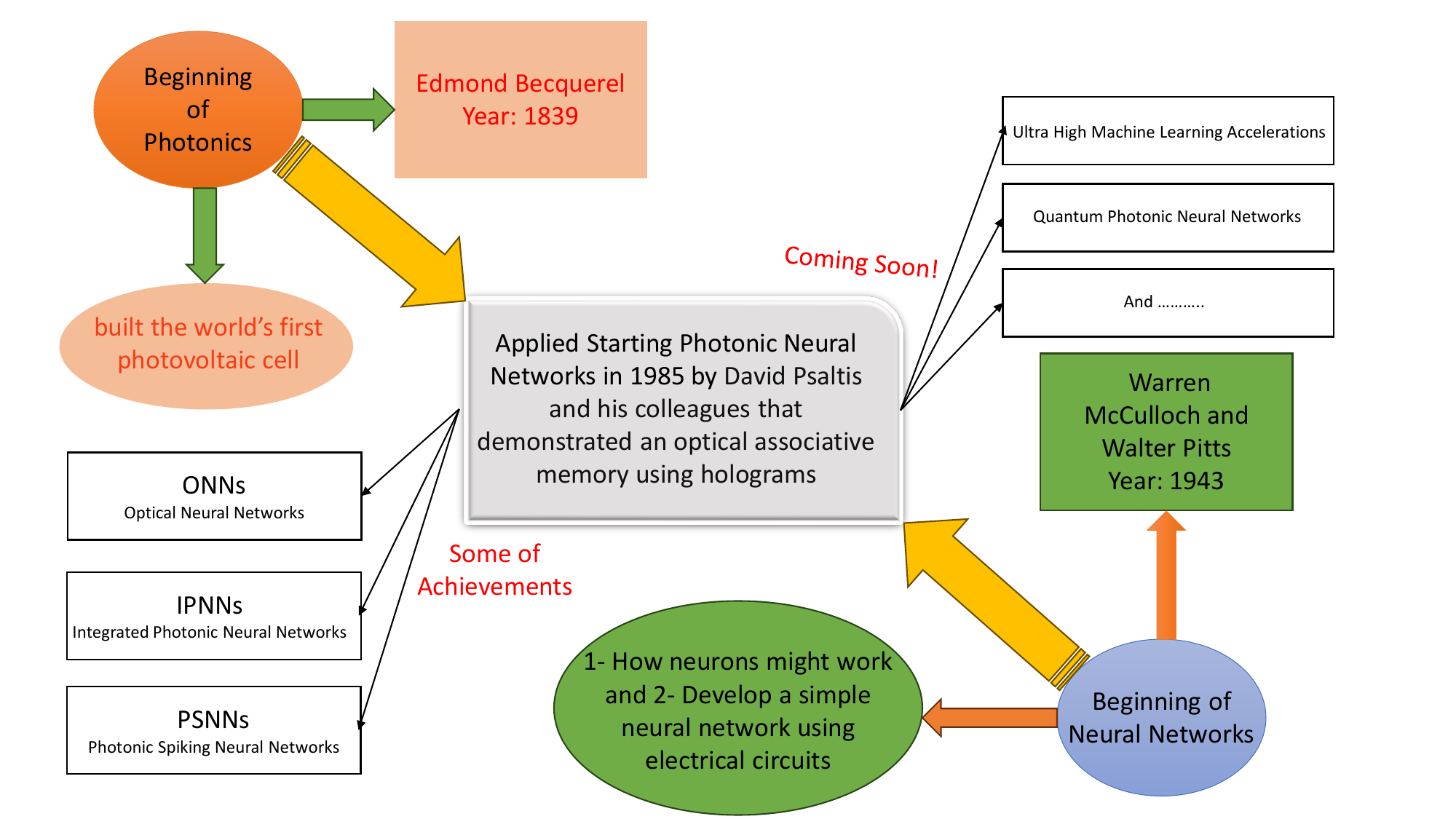}
			\caption {A brief history and outlook of Photonic Neural Networks.}
		\end{center}
	\end{figure}

	\section{Discussion}
	This presented research tries to show this note that photonics and neural networks and finally the combination of these two are important for continuing progress in doing and processing information. with photonics and other applications of photonics is still open. It has been seen that the advantages of photonic and photonic neural networks were very useful and the role of these sciences in the world is very important.
	This research has used and has studied about 80 articles but about 18 articles were main for our research. We understood that these applications are not the end of the photonic neural network's way. 
	In addition, the photonic manufacturing of the nonlinear
	function and concepts of vector-matrix are very relevant aspects. This paper also has presented the advantages and disadvantages of neural networks and compared them to electronic devices and methods. Well actually as you saw in this research we tried to show L3 cavity can be optimized by designing a neural network and we can increase the Q-factor in the L3 cavity. After this topic, we have to accept that photonics increases the speed of processing and even calculation because it is combined with neural networks, especially at short wavelengths and electronics are not able to achieve them. In many of these articles, these sentences said [11,12] although we know electronic science is very valid in all of part of the industrial and other parts of the world.
	One another aspect of this article was this question \textbf{Where are we now?} or \textbf{What are the applications of photonic neural networks?}
	In this research, we could show  \textbf{What is the photonics?}  or \textbf{ What is the neural networks?}. And then we explained \textbf{What is the relation between photonics and neural networks?}. In addition, we used some other articles that studied each of the topics separately. For example for CNN  and ONN, we used related articles of them. 
	Finally, we should say we also used and got help from many other related articles except 17 main articles. All of the other articles are referenced in the final part. Finally, we explain some of the benefits and challenges of photonic neural networks.\\
	In the final work, we stated the differences between Electronic and Photonic devices. The most important differences are about MAC (Multiply and Accumulate)/J(Energy) and also MAC/s(time).\\\\
	
	\textbf{Q3}: \textit{What are the most important challenges of Photonic Neural Network?}\\
	
	\noindent\textbf{Challenges of Photonic Neural Networks:}\\
	\noindent 1. Integration with Existing Technology: Integrating photonic components into existing electronic hardware and infrastructure can be challenging. This requires the development of hybrid systems that can efficiently interface with both photonic and electronic components.
	
	\noindent 2.Scalability: While photonic neural networks offer significant advantages in terms of speed, scaling up these systems to manage difficult tasks and large datasets can be challenging. making up photonic systems that can scale useful is an important challenge.
	
	\noindent 3.Manufacturing Complexity: Fabricating photonic devices and circuits can be complex and expensive. Achieving the required precision and consistency in manufacturing can be a barrier to widespread adoption.
	
	\noindent 4. Nonlinearity: Photonic devices typically exhibit linear behavior, while neural networks sometimes need nonlinear functions. Incorporating nonlinearity functions into photonic devices can be challenging and may require additional conditions or formulations.
	
	\noindent 5.Loss and Noise: Photons can be subject to losses and noise as they propagate through optical components and fibers. Managing and minimizing these losses and noise sources is critical to the performance of photonic neural networks.\\
	
	\noindent\textbf{Some Best Benefits of Photonic Neural Networks:}
	
	\noindent 1. High Speed: Photons usually travel at the speed of light ($3\cross 10^{8} m/s$), making photonic neural networks potentially much faster than old electronic neural networks! This can lead to main progress in processing speed.
	
	\noindent 2.Energy Efficiency: Photonic devices are often more energy-efficient than their electronic counterparts, especially for tasks that involve large-scale matrix multiplications, which are common in neural network operations.
	
	\noindent 3.Low delation: Photon-based communication can reduce delation, which is crucial for applications like real-time image and speech processing.
	
	\noindent 4. Optical Computing: Photonic neural networks can process particular tasks that are complain for electronic counterparts, such as solving differential equations or optimization problems.
	
	\noindent 5.Data Security: Photonic neural networks can offer enhanced security features since they can be less susceptible to certain types of attacks, such as electromagnetic interference.\\
	
	\textbf{Q2}:\;\textit{Although Electronics has progressed in many parts of industrial work or other things, what is the location of Photonics, and what are the challenges between Electronics and Photonics?}
	
	\begin{center}
		\begin{tabular}{|p{9cm}|p{9cm}|p{3.5cm}|}
			\hline
			\begin{center}
				\textbf{Photonic Devices}
			\end{center} & \begin{center}
				\textbf{Electronic Devices}
			\end{center} & \begin{center}
				\textbf{Feature Challenge}
			\end{center} \\
			\hline  
			\vspace{0.2cm}	\begin{center}
				Photonic devices are sufficient to insert and deliver many operations per second.
			\end{center} &\vspace{0.2cm} \begin{center}
				In the same device, Electronic couldn't see many nonlinear operations at a very large size.
			\end{center}& \begin{center}
				High efficiency at operating the variables in the matrix.
			\end{center}\\
			\hline 
			\begin{center}
				This device can operate on function and process with high efficiency and the order of $10^{16}$ (MAC/J) (All Optical).
			\end{center} & \begin{center}
				They are enabled to process about the order of $10^{12}$ (MAC/J) (Electro-Optical).
			\end{center} & \begin{center}Efficieny at nonlinear operating [7]\end{center}\\
			\hline  
			\begin{center}
				Also, this device includes very high speed at the order of $10^{12}$ for each neuron (MAC/s(second)).
			\end{center} & \begin{center}
				At this device, the rate of speed is about the order of $10^{10}$ for each neuron (MAC/s(second)).
			\end{center} &\begin{center}
				Speed compersion [7]
			\end{center} \\
			\hline
		\end{tabular}
		\captionof{table}{Compersion the features between Electronic or Electronic-Optic and Photonic devices}
	\end{center}

	\section{References}	
	[1] Zhigang Chen, and Mordechai Segev. 
	Highlighting photonics: looking
	into the next decade. (2021)
	\\
	\noindent
	[2] 
	Jia Liu1, Qiuhao Wu1, Xiubao Sui1, Qian Chen1, Guohua Gu1, Liping Wang1 and Shengcai Li2. Research progress in optical neural
	networks: theory, applications, and
	developments. (2021)\\
	\noindent
	[3] Cansu Demirkiran Furkan Eris Gongyu Wang Jonathan Elmhurst Nick Moore
	Nicholas C. Harris Ayon Basumallik Vijay Janapa Reddi Ajay Joshi Darius Bunandar
	Boston University Lightmatter Harvard University. An Electro-Photonic System
	for Accelerating Deep Neural Networks.(2021)\\
	\noindent
	[4] Cite as: APL Mater. 7, 100903 (2019); https://doi.org/10.1063/1.5109689
	Submitted: 10 May 2019 • Accepted: 19 September 2019 • Published Online: 10 October 2019
	Mario Miscuglio, Gina C. Adam, Duygu Kuzum, et al.	Roadmap on material-function mapping for
	photonic-electronic hybrid neural networks.(2019)\\
	\noindent
	[5] LORENZO DE MARINIS  , MARCO COCOCCIONI  , (Senior Member, IEEE),
	PIERO CASTOLDI, (Senior Member, IEEE),
	AND NICOLA ANDRIOLLI  , (Senior Member, IEEE). Photonic Neural Networks: A Survey. (2019)\\
	\noindent
	[6]
	J. BUENO , S. MAKTOOBI , L. FROEHLY , I. FISCHER , M. JACQUOT , L. LARGER , AND D. BRUNNER2 
	1Instituto de F´sica ı Interdisciplinar y Sistemas Complejos, IFISC (UIB-CSIC), Campus Universitat de les Illes Baleares, E-07122 Palma de Mallorca, Spain
	2FEMTO-ST Institute/Optics Department, CNRS and University Bourgogne Franche-Comté, 15B avenue des Montboucons,
	25030 Besançon Cedex, France.
	Reinforcement learning on a large scale
	photonic recurrent neural network.(2018)\\
	\noindent
	[7]
	MARIO MISCUGLIO , ARMIN MEHRABIAN , ZIBO HU , SHAIMAA I. AZZAM ,
	JONATHAN GEORGE, ALEXANDER V. KILDISHEV, MATTHEW PELTON,
	AND VOLKER J. SORGER.
	All-optical nonlinear activation function for
	photonic neural networks [Invited]. (2018)\\
	\noindent
	[8] Takashi Asano and Susumu Noda. Iterative optimization of photonic crystal
	nanocavity designs by using deep neural networks.(2019)\\
	\noindent
	[9] Pascal Stark, Folkert Horst, Roger Dangel, Jonas Weiss and Bert Jan Offrein. Opportunities for integrated photonic neural
	networks.(2020)\\
	\noindent
	[10] ALEC M. HAMMOND AND RYAN M. CAMACHO.
	Designing integrated photonic devices using
	artificial neural networks.(2019)\\
	\noindent
	[11] Xingyuan Xu,  Mengxi Tan,  Bill Corcoran,  Jiayang Wu, Andreas Boes,  Thach G. Nguyen, 
	Sai T. Chu,  Brent E. Little,  Damien G. Hicks, Roberto Morandotti,  Arnan Mitchell,  and
	David J. Moss.
	11.0 Tera-FLOP/second photonic convolutional accelerator
	for deep learning optical neural networks.\\
	\noindent
	[12] TIAN ZHANG, JIA WANG,  YIHANG DAN,  YUXIANG LANQIU, 
	JIAN DAI ,  XU HAN ,  XIAOJUAN SUN ,  AND KUN XU. Efficient training and design of photonic neural
	network through neuroevolution.(2019)\\
	\noindent
	[13] Alexander N. Tait, Thomas Ferreira de Lima, Mitchell A. Nahmias,
	Heidi B. Miller,y Hsuan-Tung Peng, Bhavin J. Shastri,y and Paul R. Prucnal
	Department of Electrical Engineering, Princeton University, Princeton, NJ 08544, USA
	(Dated: January 1, 2019). A silicon photonic modulator neuron.(2018)\\
	\noindent
	[14]
	Ryan Hamerly, Liane Bernstein, Alexander Sludds, Marin Soljačić , and Dirk Englund
	Research Laboratory of Electronics, MIT, 50 Vassar Street, Cambridge, Massachusetts 02139, USA.
	Large-Scale Optical Neural Networks Based on Photoelectric Multiplication. (2019)\\
	\noindent
	[15] Changming Wu, Heshan Yu, Seokhyeong Lee1, Ruoming Peng, Ichiro Takeuchi, and MoLi. 
	Programmable phase-change metasurfaces on
	waveguides for multimode photonic convolutional
	neural network.(2021)\\
	\noindent
	[16]
	TYLER W. HUGHES, MOMCHIL MINKOV, YU SHI, AND SHANHUI FAN
	Department of Applied Physics, Stanford University, Stanford, California 94305, USA \\Ginzton Laboratory and Department of Electrical Engineering, Stanford University, Stanford, California 94305, USA
	Corresponding author: shanhui@stanford.edu Training of photonic neural networks through
	in situ backpropagation and gradient measurement.(2018)\\
	\noindent
	[17] Mitchell A. Nahmias , Thomas Ferreira de Lima , Alexander N. Tait , Hsuan-Tung Peng ,
	Bhavin J. Shastri, Member, IEEE, and Paul R. Prucnal, Fellow, IEEE Photonic Multiply-Accumulate Operations for
	Neural Networks.(2020)\\
	\noindent
	[18] Inverse Mohammad H. tahersima , Keisuke Kojima , toshiaki Koike-Akino, Devesh Jha ,
	Bingnan Wang, Chungwei Lin, and Kieran Parsons
	Design of Integrated photonic
	power splitters. Deep Neural Network (2019)\\
	\noindent
	[19]
	Johnny Moughames, Xavier Porte,  Michael Thiel, Gwenn Ulliac,
	AND 
	Laurent Larger, Maxime Jacquot, Muamer Kadic, Daniel Brunner. Three-dimensional waveguide interconnects for
	scalable integration of photonic neural networks. (2020)\\
	\noindent
	[20] 
	Jiuxiang Gua, Zhenhua Wangb, Jason Kuenb, Lianyang Mab, Amir Shahroudyb, Bing Shuaib, Ting
	Liub, Xingxing Wangb, Gang Wangb, Jianfei Caic, Tsuhan Chenc. Recent Advances in Convolutional Neural Networks. (2017)\\
	\noindent
	[21] 
	Zewen Li, Fan Liu, Member, IEEE, Wenjie Yang, Shouheng Peng, and Jun Zhou, Senior Member, IEEE. A Survey of Convolutional Neural Networks:
	Analysis, Applications, and Prospects. (2021)\\
	\noindent
	[22] Zavareh, PH.; Safayari, A.; Bolhasani, H. BCNet: A Deep Convolutional Neural Network for Breast Cancer Grading. arXiv:2107.05037, Jul 2021.\\
	\noindent
	[23] 
	Y. Lu, M. Stegmaier, P. Nukala, M. A. Giambra, S. Ferrari, A. Busacca,
	W. H. P. Pernice, and R. Agarwal, Nano Lett. 17, 150 (2017).\\
	\noindent
	[24]
	C. Ríos, M. Stegmaier, P. Hosseini, D. Wang, T. Scherer, C. D. Wright,
	H. Bhaskaran, and W. H. P. Pernice, Nat. Photonics 9, 725 (2015).\\
	\noindent
	[25]
	C. Ríos, N. Youngblood, Z. Cheng, M. L. Gallo, W. H. P. Pernice, C. D. Wright,
	A. Sebastian, and H. Bhaskaran, Sci. Adv. 5 , eaau5759 (2019).\\
	\noindent
	[26]
	J. Zheng, A. Khanolkar, P. Xu, S. Colburn, S. Deshmukh, J. Myers, J. Frantz,
	E. Pop, J. Hendrickson, J. Doylend, N. Boechler, and A. Majumdar, Opt. Mater.
	Express 8 , 1551 (2018).\\
	\noindent
	[27]
	J. Feldmann, N. Youngblood, C. D. Wright, H. Bhaskaran, and W. H. P. Pernice,
	Nature 569, 208 (2019)\\
	\noindent
	[28] K. Wagner and D. Psaltis, “Multilayer optical learning networks, ” Appl.
	Opt. 26, 5061–5076 (1987).\\
	\noindent
	[29] C. Denz, Optical Neural Networks (Springer Vieweg, 1998).\\
	\noindent
	[30] L. Appeltant, M. C. Soriano, G. Van der Sande, J. Danckaert, S. Massar,
	J. Dambre, B. Schrauwen, C. R. Mirasso, and I. Fischer, “Information
	processing using a single dynamical node as a complex system, ” Nat.
	Commun. , 468 (2011).\\
	\noindent
	[31]
	F. Duport, B. Schneider, A. Smerieri, M. Haelterman, and S. Massar,
	“All-optical reservoir computing, ” Opt. Express 20, 22783–22795 (2012).\\
	\noindent
	[32]
	Y. Paquot, F. Duport, A. Smerieri, J. Dambre, B. Schrauwen, M.
	Haelterman, and S. Massar, “Optoelectronic reservoir computing, ” Sci.
	Rep. 2, 287 (2012).\\
	\noindent
	[33]
	L. Larger, M. C. Soriano, D. Brunner, L. Appeltant, J. M. Gutierrez, L.
	Pesquera, C. R. Mirasso, and I. Fischer, “Photonic information process-
	ing Beyond Turing: an optoelectronic implementation of reservoir
	computing, ” Opt. Express 20, 3241–3249 (2012).\\
	\noindent
	[34]
	D. Brunner, M. C. Soriano, C. R. Mirasso, and I. Fischer, “Parallel pho-
	tonic information processing at gigabyte-per-second data rates using
	transient states, ” Nat. Commun. , 1364 (2013).\\
	\noindent
	[35]
	Y. Shen, N. C. Harris, S. Skirlo, M. Prabhu, T. Baehr-Jones, M.
	Hochberg, X. Sun, S. Zhao, H. Larochelle, D. Englund, and M. Soljacic,
	“Deep learning with coherent nanophotonic circuits, ” Nat. Photonics 11,
	441–446 (2017).\\
	\noindent
	[36]
	P. Antonik, M. Haelterman, and S. Massar, “Online training for high-
	performance analog readout layers in photonic reservoir computers, ”
	Cognit. Comput. , 297–306 (2017).\\
	\noindent
	[37]
	B. J. Shastri, A. N. Tait, T. F. de Lima, M. A. Nahmias, H.-T. Peng, and P. R. Prucnal, “Principles of
	Neuromorphic Photonics,” arXiv:1801.00016 [physics] 1–37 (2018).\\
	\noindent
	[38]
	A. N. Tait, T. F. de Lima, E. Zhou, A. X. Wu, M. A. Nahmias, B. J. Shastri, and P. R. Prucnal, “Neuromorphic
	photonic networks using silicon photonic weight banks,” Sci. Rep. 7 (1), 7430 (2017).\\
	\noindent
	[39]
	M. A. Nahmias, B. J. Shastri, A. N. Tait, T. F. de Lima, and P. R. Prucnal, “Neuromorphic Photonics,” Optics
	and amp; Photonics News, OPN 29 (1), 34–41 (2018).\\
	\noindent
	[40] A. E.-J. Lim, J. Song, Q. Fang, et al., “Review of silicon photonics
	foundry efforts, ” IEEE J. Sel. Top. Quantum Electron. , vol.20,
	no. 4, pp. 405–416, Jul. 2014.\\
	\noindent
	[41] H. Zhao, S. Pinna, F. Sang, et al., “High-power indium phosphide
	photonic integrated circuits, ” IEEE J. Sel. Top. Quantum Electron. ,
	vol. 25, no. 6, pp. 1–10, Nov. 2019.\\
	\noindent
	[42] M. A. Taubenblatt, “Optical interconnects for high-performance
	computing, ” J. Light. Technol. , vol. 30, no. 4, pp. 448–457, Feb.
	2012.\\
	\noindent
	[43]
	M. J. Heck, H.-W. Chen, A. W. Fang, B. R. Koch, D. Liang, H. Park, M. N. Sysak, and J. E. Bowers, “Hybrid silicon
	photonics for optical interconnects,” IEEE J. Sel. Top. Quantum Electron. 17 (2), 333–346 (2011).\\
	\noindent
	[44]
	T. Barwicz, H. Byun, F. Gan, C. W. Holzwarth, M. A. Popovic, P. T. Rakich, M. R. Watts, E. P. Ippen, F. X. Kärtner,
	H. I. Smith, J. S. Orcutt, R. J. Ram, V. Stojanovic, O. O. Olubuyide, J. L. Hoyt, S. Spector, M. Geis, M. Grein, T.
	Lyszczarz, and J. U. Yoon, “Silicon photonics for compact, energy-eﬃcient interconnects (invited),” J. Opt. Netw.
	6 (1), 63–73 (2007).\\
	\noindent
	[45]
	J. Wang, “Chip-scale optical interconnects and optical data processing using silicon photonic devices,” Photon. Netw.
	Commun. 31 (2), 353–372 (2016).\\
	\noindent
	[46]
	A. Orieux and E. Diamanti, “Recent advances on integrated quantum communications,” J. Opt. 18 (8), 083002 (2016).\\
	\noindent
	[47]
	F. Flamini, N. Spagnolo, and F. Sciarrino, “Photonic quantum information processing: a review,” Rep. Prog. Phys.
	82 (1), 016001 (2019).\\
	\noindent
	[48]
	D. Bunandar, A. Lentine, C. Lee, H. Cai, C. M. Long, N. Boynton, N. Martinez, C. DeRose, C. Chen, M. Grein, D.
	Trotter, A. Starbuck, A. Pomerene, S. Hamilton, F. N. C. Wong, R. Camacho, P. Davids, J. Urayama, and D. Englund,
	“Metropolitan quantum key distribution with silicon photonics,” Phys. Rev. X 8 (2), 021009 (2018).\\
	\noindent
	[49]
	N. C. Harris, G. R. Steinbrecher, M. Prabhu, Y. Lahini, J. Mower, D. Bunandar, C. Chen, F. N. C. Wong, T. Baehr-Jones,
	M. Hochberg, S. Lloyd, and D. Englund, “Quantum transport simulations in a programmable nanophotonic processor;
	EP,” Nat. Photonics 11 (7), 447–452 (2017).\\
	\noindent
	[50]
	X. Qiang, X. Zhou, J. Wang, C. M. Wilkes, T. Loke, S. O’Gara, L. Kling, G. D. Marshall, R. Santagati, T. C.
	Ralph, J. B. Wang, J. L. O’Brien, M. G. Thompson, and J. C. F. Matthews, “Large-scale silicon quantum photonics
	implementing arbitrary two-qubit processing,” Nat. Photonics 12 (9), 534–539 (2018).\\
	\noindent
	[51]
	L. Chrostowski and M. Hochberg, Silicon Photonics Design: From Devices to Systems (Cambridge University, 2015).\\
	\noindent
	[52]
	W. Bogaerts and L. Chrostowski, “Silicon Photonics Circuit Design: Methods, Tools and Challenges,” Laser
	Photonics Rev. 12 (4), 1700237 (2018).\\
	\noindent
	[53]
	S. Rakheja and V. Kumar, in Quality Electronic De-
	sign (ISQED), 2012 13th International Symposium on
	(2012) pp. 732–739.\\
	\noindent
	[54] A. M. Weiner, Optics Communications 284, 3669
	(2011), Special Issue on Optical Pulse Shaping, Arbi-
	tray Waveform Generation, and Pulse Characterize-
	tion.\\
	\noindent
	[55] D. P´erez, I. Gasulla, L. Crudgington, D. J. Thomson,
	A. Z. Khokhar, K. Li, W. Cao, G. Z. Mashanovich, and
	J. Capmany, Nature Communications 8, 636 (2017).\\
	\noindent
	[56] W. Liu, M. Li, R. S. Guzzon, E. J. Norberg, J. S. Parker,
	M. Lu, L. A. Coldren, and J. Yao, Nat. Photon. 10, 190
	(2016).\\
	\noindent
	[57] T. Ferreira de Lima, B. J. Shastri, A. N. Tait, M. A.
	Nahmias, and P. R. Prucnal, Nanophotonics 6 (2017),
	10.1515/nanoph-2016-0139.\\
	\noindent
	[58] D. Brunner and I. Fischer, Optics letters 40, 3854
	(2015).\\
	\noindent
	[59] D. Psaltis and Y. Quio, Opt. Photon. News 1, 17 (1990).\\
	\noindent
	[60] J. W. Goodman, A. R. Dias, and L. M. Woody Opt.
	Lett. 2, 1 (1978).\\
	\noindent
	[61] P. Asthana, G. P. Nordin, J. Armand R. Tanguay, and
	B. K. Jenkins, Appl. Opt. 32, 1441 (1993).\\
	\noindent
	[62] M. Hill, E. E. E. Frietman, H. de Waardt, G.-D. Khoe,
	and H. Dorren, IEEE Trans. Neural Networks 13 , 1504
	(2002).\\
	\noindent
	[63] B. J. Shastri, A. N. Tait, T. F. de Lima, M. A. Nah-
	mias, H.-T. Peng, and P. R. Prucnal, arXiv:1801.00016
	(2018).\\
	\noindent
	[64] Bocker, R. P. Matrix multiplication using incoherent optical techniques. Appl.
	Opt. 13, 1670–1676 (1974).\\
	\noindent
	[65] Ríos, C. et al. In-memory computing on a photonic platform. Sci. Adv. 5 ,
	eaau5759 (2019).\\
	\noindent
	[66] Chakraborty, I., Saha, G. and Roy, K. Photonic in-memory computing primitive
	for spiking neural networks using phase-change materials. Phys. Rev. Appl. 11,
	014063 (2019).\\
	\noindent
	[67] Caulﬁeld, H. J., Kinser, J. and Rogers, S. K. Optical neural networks. Proc. IEEE
	77, 1573–1583 (1989).\\
	\noindent
	[68] Feldmann, J. et al. Parallel convolution processing using an integrated
	photonic tensor core. arXiv preprint arXiv:2002.00281 (2020).\\
	\noindent
	[69] Feldmann, J., Youngblood, N., Wright, C. D., Bhaskaran, H. and Pernice, W. H.
	P. All-optical spiking neuro synaptic networks with self-learning capabilities.
	Nature 569, 208–214 (2019).\\
	\noindent
	[70] MARIO MISCUGLIO,1 ARMIN MEHRABIAN,1 ZIBO HU,1 SHAIMAA I. AZZAM,2
	JONATHAN GEORGE,1 ALEXANDER V. KILDISHEV,2 MATTHEW PELTON,3
	AND VOLKER J. SORGER. All-optical nonlinear activation function for
	photonic neural networks (2018).\\
	\noindent
	[71] Min Gu, Yibo Dong, Haoyi Yu, Haitao Luan and Qiming Zhang. Perspective on 3D vertically-integrated photonic
	neural networks based on VCSEL arrays. (2023)\\
	\noindent
	[72] Sunil Pai , Carson Valdez , Taewon Park , Maziyar Milanizadeh , Francesco Morichetti , Andrea Melloni , Shanhui Fan , Olav Solgaard and David A. B. Miller. Power monitoring in a feedforward photonic network using two output detectors. (2023)\\
	\noindent
	[73] Sanmitra Banerjee , Mahdi Nikdast , Senior Member, IEEE, Sudeep Pasricha, and Krishnendu Chakrabarty. Pruning Coherent Integrated Photonic Neural Networks. (2023)\\
	\noindent
	[74] CANSU DEMIRKIRAN and FURKAN ERIS, GONGYU WANG, JONATHAN ELMHURST, NICK MOORE, NICHOLAS C. HARRIS, and AYON BASUMALLIK, VIJAY JANAPA REDDI, AJAY JOSHI, DARIUS BUNANDAR. An Electro-Photonic System for Accelerating Deep Neural
	Networks (2023)\\
	\noindent
	[75] Kun Liao, Tianxiang Dai, Qiuchen Yan, Xiaoyong Hu, and Qihuang Gong. Integrated Photonic Neural Networks: Opportunities and
	Challenges. (2023)
	\noindent
	[76] YUN-JHU LEE, MEHMET BERKAY ON, XIAN XIAO, ROBERTOPROIETTI, AND S. J. BEN YOO. Photonic spiking neural networks with
	event-driven femtojoule optoelectronic neuronsbased on Izhikevich-inspired model.(2023)

	\newpage
	
	\section{Authors}
	
	\begin{center}
		\begin{tabular}{|p{12.5cm}|p{5.5cm}|}
			\hline
			
			\vspace{0.1cm}
			\textbf{Mohammad Ahmadi}\newline
			
			BSc. Physics, University of Isfahan, 2018-2022.\newline\newline
			MSc. Photonics, Laser and Plasma Research Institute,\newline
			Shahid Beheshti University, Tehran, Iran, 2022-now.\newline\newline
			Fields of Interest: Quantum Imaging, Quantum Optics, Optical Designing, Quantum Information, Laser
			Biophotonic, Machine Learning\vspace{0.1cm}
			&
			\vspace{0.06cm}
			\includegraphics[width = 5.5cm , height = 6cm]{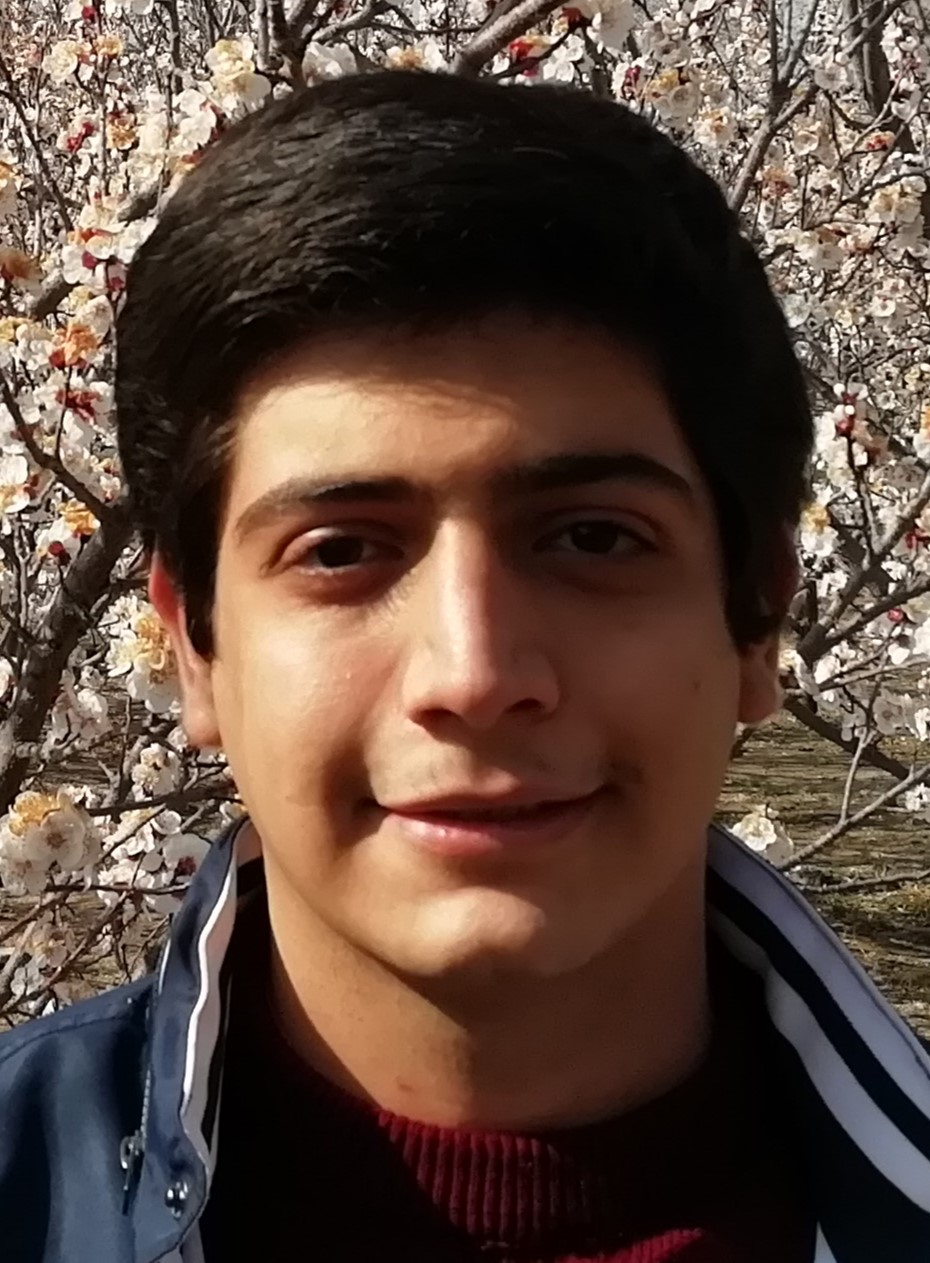}\\
			
			\hline
			\vspace{0.2cm}
			\textbf{Hamidreza Bolhasani}\newline
			Ph.D. / Researcher / Visiting Professor\newline\newline
			BSc. Physics, University of Isfahan, 2004-2008.\newline\newline
			MSc. Information Technology, Science and Research Branch,\newline
			Islamic Azad University, Tehran, Iran, 2015-2018.\newline\newline
			Ph.D. Computer Engineering, Science and Research Branch,\newline
			Islamic Azad University, Tehran, Iran, 2018-2023.\newline\newline
			Fields of Interest: Artificial Intelligence, Machine
			Learning, Deep Learning, Deep Learning Accelerators,\newline
			Computer Architecture, Neural Networks
			
			& 
			\vspace{0.02cm}
			\includegraphics[width = 5.5cm , height = 7.4cm]{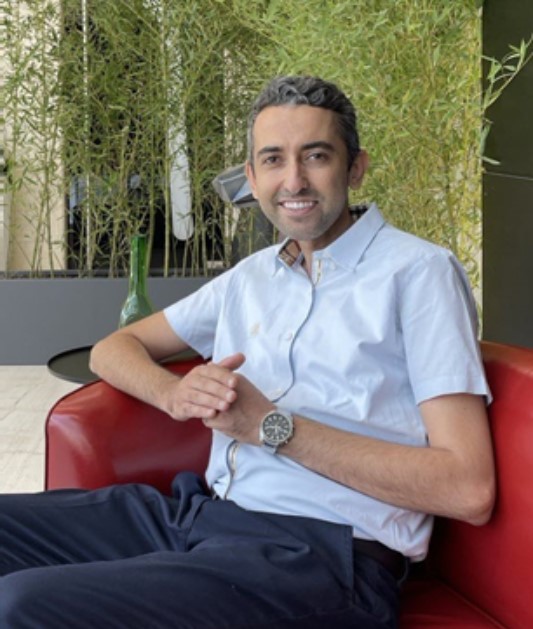}\\
			
			\hline
			
		\end{tabular} 
		
	\end{center}

\end{document}